\documentclass[a4paper]{article}

\usepackage[english]{babel}
\usepackage[utf8x]{inputenc}
\usepackage[T1]{fontenc}
\usepackage{booktabs}
\usepackage{multirow}
\usepackage{siunitx}
\usepackage{makecell}
\usepackage[export]{adjustbox}
\usepackage{soul}
\usepackage{pdflscape}
\usepackage{lscape}
\usepackage{longtable}

\usepackage[a4paper,top=3cm,bottom=2cm,left=3cm,right=3cm,marginparwidth=1.75cm]{geometry}

\usepackage{amsmath}
\usepackage{amssymb}
\usepackage{graphicx}
\usepackage[colorlinks=true, allcolors=blue]{hyperref}
\usepackage{setspace}
\usepackage[numbers,sort]{natbib}
\usepackage{appendix}
\usepackage{threeparttable}
\usepackage{environ}
\usepackage{threeparttablex}
\usepackage{eurosym}
\usepackage{textcomp,mathcomp}

\title{Machine learning for potion development at Hogwarts}
\author{Christoph F. Kurz$^{1,2}$, Adriana N. König$^{1,2}$}
\date{\small{$^1$Institute of Health Economics and Health Care Management, Helmholtz Zentrum München, Ingolstädter Landstraße 1, 85764 Neuherberg, Germany}\\[2ex]%
   $^2$Munich School of Management and Munich Center of Health Sciences, Ludwig-Maximilians-Universität München, Geschwister-Scholl-Platz 1, 80539 Munich, Germany\\[2ex]%
    \normalsize{\today}
}

\begin{document}

\maketitle

\begin{abstract}
\noindent \textbf{Objective}: To determine whether machine learning methods can generate useful potion recipes for research and teaching at Hogwarts School of Witchcraft and Wizardry. \\
\textbf{Design}: Using deep neural networks to classify generated recipes into a standard drug classification system. \\
\textbf{Setting}: Hogwarts School of Witchcraft and Wizardry. \\
\textbf{Data sources}: 72 potion recipes from the Hogwarts curriculum, extracted from the Harry Potter Wiki.  \\ 
\textbf{Results}: Most generated recipes fall into the categories of psychoanaleptics and dermatologicals. The number of recipes predicted for each category reflected the number of training recipes. Predicted probabilities were often above 90\% but some recipes were classified into 2 or more categories with similar probabilities which complicates anticipating the predicted effects. \\
\textbf{Conclusions}: Machine learning powered methods are able to generate potentially useful potion recipes for teaching and research at Hogwarts. This corresponds to similar efforts in the non-magical world where such methods have been applied to identify potentially effective drug combinations.
\end{abstract}

\onehalfspacing

\section*{Introduction}
Potions are a required subject for students at Hogwarts School of Witchcraft and Wizardry from the first to the fifth year \citep{rowling1997harry}. 
They are optional to students in their sixth and seventh-years if they achieved a high score on their Ordinary Wizarding Level exam \citep{rowling2005harry}.
Potions classes are considered to be among the most difficult lessons at Hogwarts because the nuances of timing, ageing, bottling, and stirring techniques are difficult to acquire even with the guidance of experienced teachers such as Professor Severus Snape. 
Brewing potions requires glass vials, weighting scales and a cauldron. Ingredients range from plants such as belladonna and shrivelfig to magical components such as unicorn hair or fairy wings. The brewing process often requires some degree of wand work to complete \citep{rowling1997harry}.

Potions can be used as medicines, antidotes, poisons, or to provide the drinker with any magical effect ranging from increased strength to flame immunity. They are not always consumed by drinking them; some, like the Regeneration potion, might be applied by physical touch or have an effect merely by being created \citep{rowling2000harry}.
Certain magical effects can only be achieved through the use of potions.
Some potions mimic the effects of spells and charms, but a few (such as the Polyjuice Potion, a potion that allows the drinker to take the form of someone else, and Felix Felicis, a luck potion) have effects that can not be achieved in any other way \citep{rowling1998harry,rowling2005harry}.

Because brewing is so difficult and the smallest deviations from the recipe can have serious consequences, there are countless reports of accidents and undesirable side effects happening during class at Hogwarts \citep{rowling1997harry,rowling1998harry,rowling1999harry}. For example, in 1992, there was a well-documented case of the student Neville Longbottom who, while improperly brewing the Cure for Boils potion, infected himself with red boils all over his body \citep{rowling1997harry}. 
Nevertheless, some deviations from instructions have proven successful for Harry Potter in his fifth year at Hogwarts \citep{rowling2005harry}.
Accurately following the brewing instructions is already difficult, but the discovery and development of new potions is an even more complex and dangerous process.

Recent advances in the field of artificial intelligence (AI) have led to increased interest in the use of machine learning approaches within the pharmaceutical industry. Advances in new algorithms, such as deep neural networks, demonstrated its utility 
in addressing diverse problems in drug discovery such as bioactivity prediction or novel molecular designs \citep{chen2018rise,vamathevan2019applications}.

In this work, we explore the usefulness of machine learning for generating recipes for magic potions. For this, we randomly generated new magic potion recipes with various ingredients and predicted their most likely effect using an artificial neural network.

\section*{Methods}
We collected the recipes for all known potions from the Harry Potter Wiki \citep{hpwiki2022} and classified them according to the Anatomical Therapeutic Chemical (ATC) classification system \citep{who2022atc} in one of the following categories: anesthetics; antiinfectives for systemic use; antiparasitic products, insecticides and repellents; dermatologicals;  musculo-skeletal system; psychoanaleptics; psycholeptics; respiratory system; sensory organs; and various. These categories represent the first and second level of the ATC classification which describes pharmacological or therapeutic subgroups \citep{who2022atc}.
Recipes in the musculo-skeletal system category include, for example, the pompion potion that temporarily turns the drinkers head into a pumpkin, or the skelegro potion that regrows bones.
Dermatologicals are, among others, potions that make your skin immune to fire, grow hair or cure boils. 
Recipes in the psychoanaleptics category include, for example, the forgetfulness potion which causes memory loss in the drinker. Others are the wit-sharpening potion which improves clear thinking or the befuddlement draught that provokes belligerence and recklessness in the drinker.
The various category contains several antidotes as well as potions that boost spell-casting, such as the exstimulo potion.

We additionally added a category for poisons because many recipes fall into it. However, poisons are not associated with an ATC code for obvious reasons. Usually, administered drugs aim at improving and not deteriorating an individual's health.
See Table~\ref{tab:potioncount} for an overview of the number of recipes in each of the 11 categories. In total, the training set contained 72 recipes. Each recipe includes instructions for adding ingredients and brewing.
We then generated 10,000 new potion recipes by randomly picking between 3 to 8 single ingredients (e.g., ``Add 4 horned slugs to your cauldron'') and mixing instructions (e.g., ``Stir 5 times, clockwise.''). 
We used a custom BioBERT neural network \citep{biobert2019} for predicting the class of potion for each newly generated recipe.
This language model has been pre-trained on large-scale biomedical corpora comprising over 18 billion words from PubMed abstracts and full-text articles. We fine-tuned the model to all known Hogwarts potion recipes so it would input a recipe and output the probabilities of belonging to each of the 11 classes. The top probability is the most likely effect. This method is often referred to as feature extraction transfer learning \citep{goodfellow2016deep}. All computations were done in Mathematica 13 \citep{Mathematica}. Code and data are available on our GitHub page \citep{kurz2022github}.

\begin{table}[!ht]
    \centering
    \begin{tabular}{lrr}
    \toprule
        Category & ATC Code & Count  \\ \midrule
        Anesthetics & N01 & 2 \\
        Antiinfectives for systemic use & J & 4 \\
        Antiparasitic products, insecticides and repellants & P & 2 \\
        Dermatologicals & D & 8  \\ 
        Musco-sceletal system & M & 12 \\
        Poison & -- & 11 \\
        Psychoanaleptics & N06 & 21  \\
        Psycholeptics & N05 & 2 \\
        Respiratory system & R & 1 \\
        Sensory organs & S & 1 \\
        Various & V & 8 \\
    \end{tabular}
    \caption{Number of recipes and ATC codes for each of the 11 categories in the training data.}
    \label{tab:potioncount}
\end{table}

\section*{Results}
\begin{figure}[ht]
	\centering
	\includegraphics[width=0.6\textwidth]{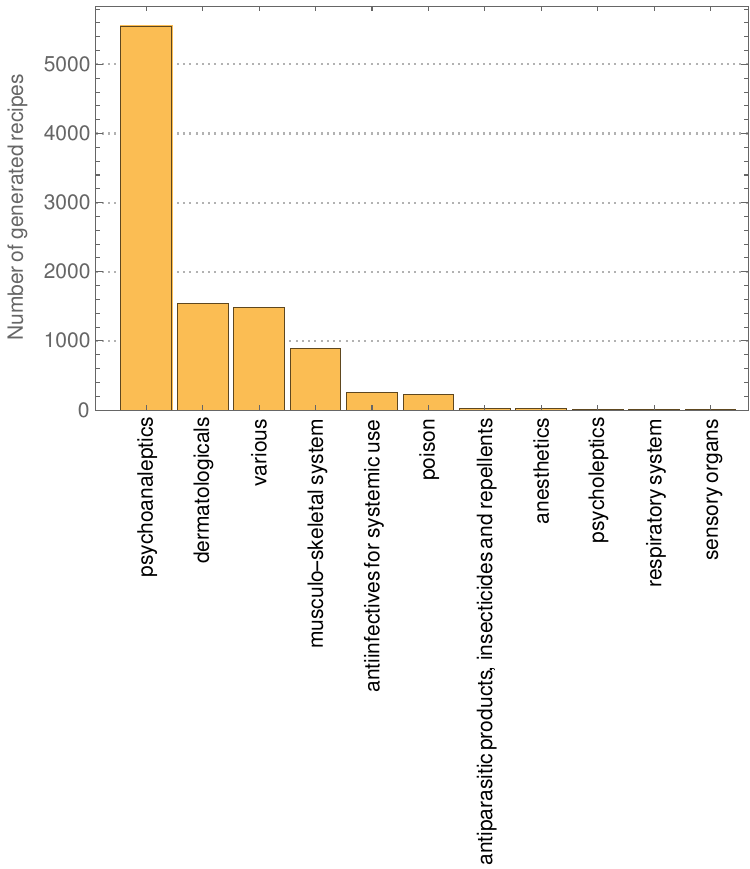}
	\caption{Predicted potion categories of the 10,000 randomly generated recipes.}\label{fig:res1}
\end{figure}

\begin{figure}[ht]
	\centering
	\includegraphics[width=0.6\textwidth]{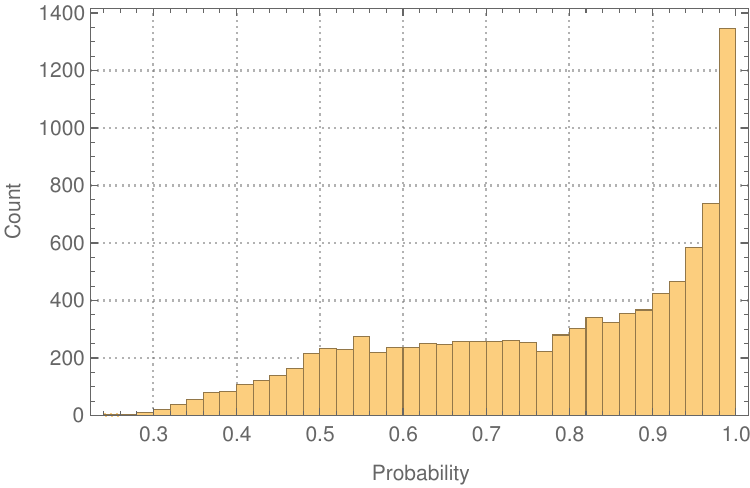}
	\caption{Count histogram of the predicted probabilities of belonging to one of the 11 ATC categories (including poisons). Only the top probability for each generated recipe is shown.}\label{fig:predhist}
\end{figure}

Figure~\ref{fig:res1} shows the number of predicted potion recipes in each category. Most of the 10,000 generated recipes fall into the psychoanaleptics category ($n=5549$), followed by the dermatologicals category ($n=1539$) and the various category ($n=1487$). 225 recipes fall into the newly added poison category.
In contrast, only 3 psycholeptics, 3 respiratory systems, and 1 sensory organs recipes were generated. This corresponds to the number of available training recipes. All generated recipes differed from the training set of recipes.
Our BioBERT model was generally confident in its predictions. Predicted probabilities of belonging to a certain ATC category were often above 90\%, see Figure~\ref{fig:predhist}.

\newpage

For example, Table~\ref{tab:psycho} shows a generated recipe where its predicted effect is in the psychoanaleptics category with a probability of 99.9\%. In contrast, the effects of some recipes are difficult to predict for our model. Table~\ref{tab:multiclass} shows a generated recipe that could be dermatological with 58.4\% probability, be a psychoanaleptic with 10\% probability, or an antiinfective for systemic use with 24.1\% probability.

\begin{table}[ht]
    \centering
    \begin{tabular}{p{0.6\linewidth}}
      Add Baneberry. Add 2 bundles of knotgrass to the cauldron. Add \
Dogbane. Add syrup of hellebore until the potion turns turquoise. Add \
a sprig of Peppermint to counteract side-effects. Add Honey water \
until it turns back to a turquoise colour. Stir four times \
anti-clockwise. Add the Infusion of Wormwood. 
    \end{tabular}
    \caption{Generated recipe that is predicted to work as a psychoanaleptic with 99.9\% probability.}
    \label{tab:psycho}
\end{table}

\begin{table}[ht]
    \centering
    \begin{tabular}{p{0.6\linewidth}}
      Add Stewed Mandrake. Add Wormwood. Add a dash of Flobberworm Mucus \
and stir vigorously. Leave to brew and return in 8 hours (Copper), 14 \
hours (Brass), or 23 hours (Pewter). Shake and add wormwood until the \
potion turns green. Slice bursting mushrooms with knife, add to \
cauldron and stir clockwise until potion turns blue. 
    \end{tabular}
    \caption{Generated recipe that is predicted to work as a dermatological with 58.4\% probability, as a psychoanaleptic with 10\% probability, and as an antiinfective for systemic use with 24.1\% probability.}
    \label{tab:multiclass}
\end{table}

\section*{Discussion}
Our findings suggest that AI powered methods are able to generate potentially useful potion recipes for teaching and research at Hogwarts School of Witchcraft and Wizardry. We were able to produce many previously unknown combinations of ingredients and stirring instructions that were predicted to belong to a specific bioactivity class with high probability. 
Previously, AI methods have also been used to identify potentially effective drug combinations \citep{guvencc2021machine,wu2022machine}. In the magical world, our research could be extended to not only detect new combinations of ingredients but also new combinations of potions.
Apart from new effective combinations, AI methods could also be applied to identify potentially harmful drug combinations \citep{guvencc2021machine}. This complements our predictions in the category for poisons and could be extended to harmful potion combinations.

Still, our results are not without limitations. In general, AI models need very large training sets. We only had a set of 72 recipes available for training. For this reason, we used a model that has been pre-trained on large medical corpora. Still, potions belonging to the same ATC category often have very different effects. For example, both the Babbling Potion, a potion that causes the drinker to babble nonsense, and Baruffio's Brain Elixir, a potion that increases the drinker's brain power, are part of the nervous system category. This makes it extremely difficult to predict specific potion effects, other than the organ or system on which they act.

Furthermore, our AI approach for drug discovery and potion generation could potentially be misused.
For example, \citet{urbina2022dual} trained an AI model that generated new molecules that were predicted to be more toxic than publicly known chemical warfare agents. In this sense, machine learning could be used to support the Dark Arts. The Dark Arts refer to spells and actions that could harm others, such as powerful curses, as well as brewing dark potions and breeding dark creatures. Our approach could lead to the discovery of new spells and potions that would enable Dark Wizards or Witches becoming even more powerful than Lord Voldemort, considered to have been the most capable and dangerous practitioner of the Dark Arts of all time \citep{rowling2005harry}.

At last, two muggles with (presumably) no magical abilities performed the study. Thus, it is difficult to assess the validity and classification quality of the generated recipes.

\clearpage

\subsection*{Competing interests statement}
All authors have completed the Unified Competing Interest form (available on request from the corresponding author) and declare: no support from any organisation for the submitted work; no financial relationships with any organisations that might have an interest in the submitted work in the previous three years, no other relationships or activities that could appear to have influenced the submitted work.

\subsection*{Details of contributors}
CFK and ANK collected the data and wrote the manuscript. CFK analysed the data. ANK is the guarantor.

\subsection*{Transparency declaration}
The lead author (the manuscript's guarantor) affirms that this manuscript is an honest, accurate, and transparent account of the study being reported; that no important aspects of the study have been omitted; and that any discrepancies from the study as planned have been explained.

\subsection*{Ethical approval}
The data were obtained from publicly available data sources.

\subsection*{Details of funding}
No funding was received.

\subsection*{Patient and public involvement statement}
Not applicable.


\bibliographystyle{vancouver-authoryear}
\bibliography{biblio}

\end{document}